\documentclass[letterpaper, 10 pt, conference]{ieeeconf}  % Comment this line out if you need a4paper

\IEEEoverridecommandlockouts                              % This command is only needed if 
\usepackage{amsmath,amsfonts}
\usepackage{algpseudocode} % not 'algorithmic' %added by mostafa
\usepackage{algorithm}
\usepackage{array}
\usepackage[caption=false,font=normalsize,labelfont=sf,textfont=sf]{subfig}
\usepackage{textcomp}
\usepackage{stfloats}
\usepackage{url}
\usepackage{verbatim}
\usepackage{graphicx}
\usepackage{cite}
\usepackage{xcolor} %added by mostafa
\usepackage{siunitx} %added by Harshit
\usepackage{subcaption} %added by prof libby
\usepackage{xcolor}

\title{\LARGE \bf
Robust Silicone Pour Casting and Sensor Embedding Procedures for Soft Robotic Actuators
}

\author{Harshit Thakker$^{1}$,~\IEEEmembership{Graduate Student Member,~IEEE,}, Paul Dela Cruz$^{1}$,~\IEEEmembership{Graduate Student Member,~IEEE,},\\
Mostafa Mo. Massoud$^{1}$,~\IEEEmembership{Graduate Student Member,~IEEE,}
and Jacqueline Libby$^{2}$,~\IEEEmembership{Member,~IEEE}% <-this % stops a space
\thanks{This work was supported by the US National Science Foundation under Grant 2502197. (\emph{Corresponding author: Jacqueline Libby})}%
\thanks{$^{1}$Harshit Thakker, Paul Dela Cruz and Mostafa Mo. Massoud are with the Department of Mechanical Engineering, Stevens Institute of Technology, 1 Castle Point Terrace, Hoboken, NJ, 07030, USA
        {\tt\small hthakker@stevens.edu,pdelacr1@stevens.edu, mmassoud@stevens.edu}}%
\thanks{$^{2}$Jacqueline Libby is with Faculty of Mechanical Engineering, Stevens Institute of Technology, 1 Castle Point Terrace, Hoboken, NJ, 07030, USA
        {\tt\small jlibby@stevens.edu}}%
}

\begin{document}

\maketitle
\thispagestyle{empty}
\pagestyle{empty}

%----------------------------------------------------------------
\begin{abstract}
Soft robots are well-suited for applications such as rehabilitation and surgery that require adaptable and safe interaction with their environment.
%Applications include wearables for rehabilitation, surgery, and manufacturing pipelines that require handling of fragile objects.
%Wearable robots, surgical robots, and manufacturing plants that require handling of fragile objects are applications that can benefit from incorporating soft robots.
However, the challenges of reproducible and scalable fabrication of soft robots limit their real-world deployment.
Various fabrication methods have been introduced, but many are labor-intensive and prone to human error. 
Therefore, traditional two-part pour casting remains an attractive option.
%The automated knitting and 3D printing technologies have great potential but are still not widely adopted for mass production.
This paper presents procedures for robust, repeatable, and scalable fabrication of soft pneumatic actuators using two-part pour casting.
The presented methods prevent internal cavity clogging and ensure air-tight sealing.
%The method presented here addresses the clogging problem in two-part pour casting techniques.
%An airtight silicone pneumatic actuator is achieved using this technique. 
Additionally, a robust sensor embedding procedure for thin-film flex sensors is presented, which allows for accurate and repeatable data acquisition.
%1-2 sentences on
%  FEM
%  sensor calibration (flex sensor and image processing, real-world experiments)
%  pressure control, angle response
%  comparing FEM and angle response
% Predictions from Finite Element Modeling (FEM) are validated against real-world experiments, collecting image processing and data from the SpectraFlex Flex sensor.
%original sentence
%Predictions from Finite Element Modeling (FEM) at lower pressures are comparable to the real-world performance of the actuator and deviate moderately at higher pressures.
%which is validated by image processing and data from the SpectraFlex Flex sensor.
%modification 1
%Predictions from Finite Element Modeling (FEM) are comparable to the real-world performance of the actuator at lower pressures and deviate moderately at higher pressures.
%which is validated by image processing and data from the SpectraFlex Flex sensor.
%modification 2: FEM
%Predictions from Finite Element Modeling (FEM), at lower pressures, are comparable to the real-world performance of the actuator, and at higher pressures, deviate moderately.
Finite Element Modeling (FEM) of the soft actuator is performed to analyze stress and deformation from internal pressure loadings.
%FEM bending angle predictions are compared to real-world actuator performance.
%which is validated by image processing and data from the SpectraFlex Flex sensor.
%The performance of the actuator was validated by pressure control tests for staircase and sinusoidal inputs.
%Sensor Calibration
Pneumatic actuation experiments with PID pressure control are performed.
%After fabrication, an actuator is pneumatically actuated with PID control of internal air pressure.
Automated image processing is used to calibrate the embedded flex sensor to bending angle measurements.
%as it is pressurized over the entire bending range.
%Flex sensor calibration is performed by fitting a third-order polynomial to the data of angles obtained by image processing and sensor voltage.
%The fit polynomial is used for angle tracking for pressure control tests.
%Pressure Control Test.
Staircase and sinusoidal profile actuation experiments validate the performance of the fabricated actuator. 
Angle response experiments for the staircase input show repeatable performance, and the sinusoidal input shows a small amount of hysteresis consistent with viscoelastic response to pneumatic actuation of soft actuators.
Simulated and real-world bending angles show comparable response.
%, deviating somewhat at higher pressures.
The methods and experiments together provide a repeatable and robust fabrication procedure, validated across two operators and 24 successful fabrications, along with benchmark simulations and experimental testing.
These methods and benchmarks will enable
%The methods and experiments together benchmark
% A replicable and scalable fabrication and tens procedure is achieved, enabling 
more widespread adoption of soft robotics in real-world applications.

\end{abstract}
%----------------------------------------------------------------

%----------------------------------------------------------------
\section{INTRODUCTION}
%----------------------------------------------------------------
Soft robots have relevance across diverse applications ranging from rehabilitation devices to surgical robots to surgical implants to handling fragile components in manufacturing plants~\cite{yin2024wearable, zhang2025soft, wang2024pioneering}. Soaring healthcare costs provide a unique opportunity for the application of robotic systems in the rehabilitation space, given the repetitive nature of exercise over long time durations~\cite{gower2025breaking, laut2016present}. Rigid-body robotic devices for patient rehabilitation are less comfortable, and misalignment can cause force application in non-ideal regions~\cite{pan2022soft,morris2023state,bessler2021safety}.  Soft robots can have infinite degrees of freedom and the ability to passively adapt to their environment, making them safer for human-robot interaction applications~\cite{arnold2017tactile, xiao2024design, prabhakar2025soft}. 

There are a wide range of actuation techniques for soft robots, including fluidic/pneumatic, electric, electrohydrodynamic, electrothermal, magnetic, and chemical actuation techniques~\cite{li2022soft}. For rehabilitation applications, pneumatic soft actuators are a prevalent choice since they are cost-effective and can apply high forces~\cite{robertson2017soft}. Furthermore, they allow the design of complex elastomeric shapes capable of physical interaction with humans and delicate objects, with force distribution over the entire surface of the object~\cite{ilievski2011soft,whitesides2018soft}.
The robotics industry is heavily dominated by rigid-link robots, for which there are decades of knowledge in design, manufacturing, and control.
One of the biggest challenges for the soft robotics field to gain traction is to have robust and repeatable fabrication techniques.

Traditional soft pneumatic actuators consist of simple bladders with motion behaviors encoded through fiber-reinforcements~\cite{polygerinos2015modeling, habibian2022evaluation,kokubu2024development}. Fiber-reinforced pneumatic actuators have a simple tubular geometry which is easy to fabricate. However, the wrapping of the fiber is not trivial and requires manual labor. Furthermore, the single-bladder design makes them bulky. 
A knit textile-based actuator was introduced in~\cite{cappello2018assisting,ge2020design,luo2022digital}, which has shown potential in hand rehabilitation.
%Two separate bladders for flexion and extension are sandwiched between textile layers. The knitting technique of the fabrics controls the bending behavior of the actuators, with a stretchier fabric used for the top layer and a less stretchier fabric for the other two layers. 
Automated knitting replaces manual fiber wrapping, requiring less labor-intensive fabrication. However, the actuators remain bulky since a single bladder is used for underactuated unidirectional motion.
%The specialized textile knitting technique is still in its infancy.

%More recently, 3D printing has been explored as a potential fabrication technique for soft robots and
3D printing shows great potential for soft robotic fabrication, especially for creating intricate internal geometries.
%, especially with 
Fused Deposition modeling (FDM) with thermoplastic elastomers (TPE)~\cite{Zhai2023-SciRobot,Zhai2025-AdvIntellSyst,hu2020bioinspired} and vat polymerization with elastic resins
%techniques such as SLA, DLS, DLP
have been employed~\cite{wallin20203d,ge2022shaping};
however, obtaining air-tight prints is challenging due to layer delamination, material choices are limited, and support material for internal cavities is not possible.
FDM of TPEs requires complex cooling processes and printer modifications, and vat polymerization requires complex post-processing. 
Direct Ink Writing (DIW) is a technique where two-part silicones are 3D printed with a static mixer~\cite{Yirmibesoglu2018-RoboSoft}, but more research is necessary to prove effectiveness for airtight soft actuator applications.
%but is still in its infancy. The challenges with DIW are layer adhesion issues and air leaks.

Hence, traditional casting techniques remain a prevalent method for fabricating soft pneumatic actuators. Sacrificial core and lost wax casting techniques allow the actuator body to be fabricated from a single cast, but pose significant challenges.
%are single-step casting techniques that are especially advantageous to use for actuators with more complex .
Lost wax casting is a laborious process with pre- and post- processing steps, including casting of delicate wax cores and multiple rounds of cleaning out the wax.
%however, the cores can be  downside of this technique is that if the core has defects, then it propagates to the actuator; removing the core is a laborious process
~\cite{greer2021soluble,silva2024integrated,cervera2024lost}.
Low-volume cores (LVCs) are less laborious to remove as they can be pulled out; however, the fabrication of LVCs requires manual shaping of heat-bonded polypropylene sheets, which is prone to human error. Furthermore, the walls of the soft actuator body have to be tough enough to avoid damage when the core is pulled out, and the complexity of internal geometries is limited~\cite{yu2024low}.

\begin{figure}[ht!]
\centering
\includegraphics[width=.9\linewidth]{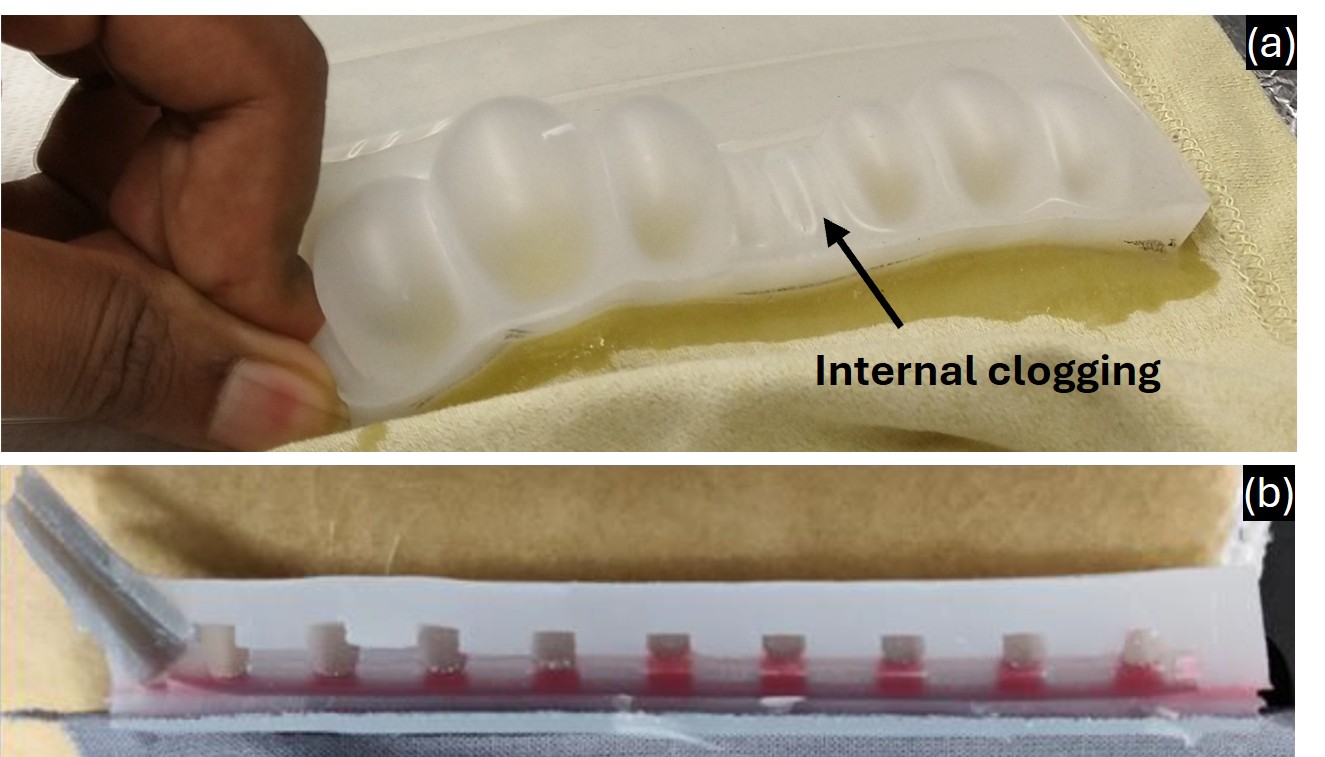}
\caption{Demonstration of uncontrolled sealing layer fabrication causing clogging. (a) The soft actuator is inflated, with clogging in the center. (b) Sectional view after dissection shows sealing layer (pink pigment), rising up along the walls due to capillary action.}
\label{fig.cloggingRealWorld}
\end{figure}

Two-part pour casting is a traditional soft actuator fabrication technique, where two open parts are cast and cured, and then sealed together to enclose the internal cavities of the pneumatic actuator.
While this technique
%two-part pour casting with a sealing layer to join the two parts
has limitations on the geometrical complexity of the sealed section, it remains a faster, cleaner, and more scalable method than other alternatives, making it widely adopted for pneumatic soft actuator casting~\cite{Singh2023-AIM,lv2025design}.
However, fabrication of the sealing layer remains a challenge.
%The fabrication of the sealing layer, which joints the
Sil-Poxy has been used to seal the two parts~\cite{heung2024quasi}; however, this introduces material non-homogeneity and can cause air leaks at the sealing.
%An alternative approach is explored to have 
An actuating core separate from the main body has been proposed to tackle air leaks~\cite{li2020high}, but assembling the actuating core into the main body is not trivial.
%, and improper assembly can lead to unexpected behaviors.
Additionally, the sealing layer can cause clogging of the internal cavities in soft actuators~\cite {low2016rod}.
The result of uncontrolled fabrication of the sealing layer is demonstrated by our experiments shown in Fig.~\ref{fig.cloggingRealWorld}.
Clogging results in non-uniform air chamber
expansion (Fig.~\ref{fig.cloggingRealWorld}a), leading to unwanted and unrepeatable bending behavior. 
Fig.~\ref{fig.cloggingRealWorld}b shows a sectional view of an actuator after it is dissected, which shows that the cause of the clogging is capillary action of the sealing layer (pigmented in pink), rising up along the internal walls.
%, and it has been observed that soft actuators fabricated using this technique usually fail at the sealing layers~\cite{torzini2024characterization}.
Robust and repeatable techniques are therefore needed for the fabrication of the sealing layers between two-part soft actuator casts. 

In addition to the sealing layer, the embedding of sensors in soft actuators is a nontrivial fabrication process, which is important for data acquisition and performance tracking.
Commercially-available flexible thin-film sensors can be embedded in flat regions of soft actuators; however, problems can arise such as 
%they need to be embedded such that no external noise is introduced during actuation.
sensor exposure to internal air flow during actuation
and sensor damage due to heat in fabrication processes~\cite{koivikko2022integrated,yuen2018strain}.
%Non-embedded sensor adherence, such as glue or stitching, can lead to non-robust attachments and sensor measurements~\cite{aljaber2024optimal}. 

In this paper, we present a two-part silicone pour casting technique including:
\begin{itemize}
    \item A repeatable procedure for fabricating robust, airtight sealing layers that do not clog the internal cavity channels. 
    \item A repeatable procedure for embedding flexible thin-film sensors without introducing noise to the system or damaging the sensor.
    \item The embedding of a fabric strain-limiting layer that extends outside of the actuator for integration with a wearable exosuit or other soft form factors.
\end{itemize}
We present these techniques on a typical fast pneumatic network (fPN) actuator~\cite{mosadegh2014pneumatic}, a common building block in soft robotics applications. Furthermore, we use SORTA-Clear 40 silicone (Smooth-On, Inc.), which is particularly challenging to seal but attractive due to its transparency, food-grade quality, high shore hardness, and low-hazard processes.

%------------------------------------------------------------------
%reference materials
%------------------------------------------------------------------
%CAD
%	silicone/CAD/newSealingLayer/
%		new updates of lid and frame from Cal's design
%	silicone/CAD/fastPneunetActuators/fastPneunetMolds/fpnFlexSensor2Mold/
%tutorials
%	harshit's tutorial
%		silicone/tutorials/sensorEmbedding/spectraFlexSensor/
%	older tutorials
%		silicone/tutorials/fpnActuator/fpnNewSealing

%------------------------------------------------------------------
\section{Methods}\label{sec.methods}
%------------------------------------------------------------------
%Sensor aligner
%New sealing layer to avoid clogging
%Using fabric instead of paper to integrate into wearable exosuit
%Air adaptor
%non-planar top
%assembly clamp for larger system
%external cuts for fpn, differentiating from spn, allows for less damage of silicone due to bending deformation, as well as faster actuation.

%One of the main reasons to use fabric as opposed to paper is that it allows direct embedding of the actuator into a piece of clothing, which opens up a large number of avenues for soft wearable rehabilitation robotic devices.

%Brief overview of the subsections, maybe here is where you say all the stuff about why things are important 
%notes on CAD
%  slide 49, started how we want the cad to look for sealing layer
%  do this as well for sensor embedding
%  any time there's a detail the CAD dimensions that had to be designed or dedesigned
%    good time to make this closeup image, like on slide 49, and then talk to the image

%to write this:
% cropping video screenshots and make them figures in paper
% write section, talking to these figures, where you explain the whole process
% as you're writing, you think of details that need a CAD
% make the CAD and add it in

\begin{figure}[ht!]
\centering
\includegraphics[width=1\linewidth]{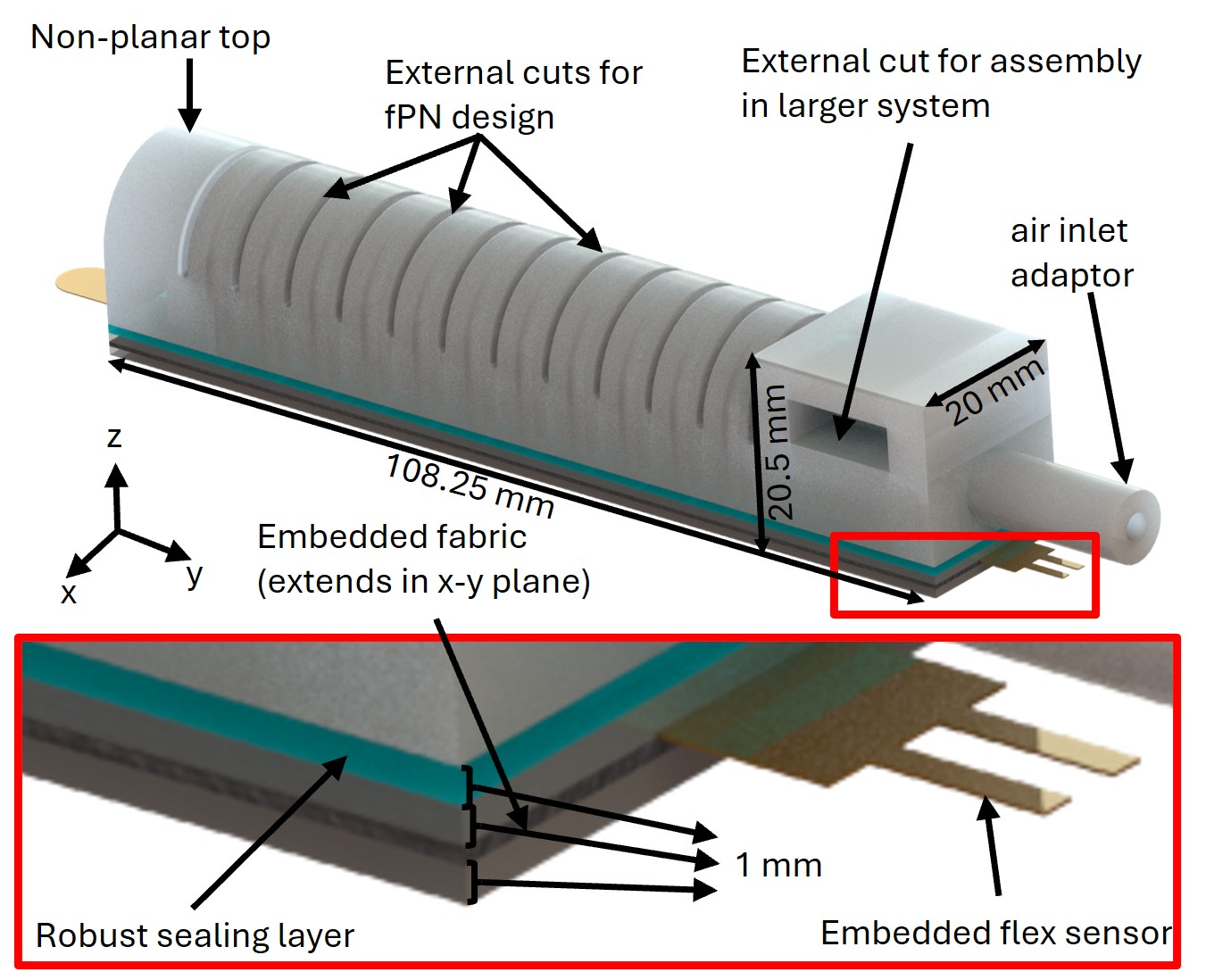}
\caption{Features of the fPN actuator.}
\label{fig.methodsIntro}
\end{figure}

In this section, we present fabrication techniques to enable a repeatable, scalable process for manufacturing robust, fast pneumatic network actuators (fPNs) with embedded flex sensors using silicone pour casting. Fig.~\ref{fig.methodsIntro} shows a model of a finished actuator, which is fabricated using a two-part casting process and then joined with a sealing process.
%The molds are made of 3D printed PLA and are reusable. 
A 95mm SpectraFlex Flex Sensor (Spectra Symbol Corp.) is embedded, which is used to track the bending angle.
The other features of this fPN include
a robust sealing layer that plays a vital role in ensuring no air leaks occur as well as an embedded fabric that can be extended in the x-y plane for integration with other soft form factors.
The fabric used is a classic cotton quilting fabric (Kona cotton, Robert Kaufman Fabrics).
Additional features include
%a robust air inlet adapter to address the air leak issues commonly observed in such actuators,
an external cut in the silicone to allow for assembled clamps, and a rounded top made possible through an inverted fabrication process of the main body.   

Section~\ref{subsec.base} discusses fabrication of the base layer, including the embedding of the flex sensor and the fabric layer. Section~\ref{subsec.main} discusses fabrication of the main body, including external cuts for fPN functionality and integration with a larger system. Section~\ref{subsec.sealing} discusses the repeatable process for fabrication of a robust air-tight sealing layer to seal the base and main body.

\subsection{Fabrication of Base with Embedded Fabric and Sensor}\label{subsec.base}
%sensor embedding (and base layer fabrication)
%	why sensor aligner is important
%	  why offsets help
%	  getting rid of human error
%	  how sensor aligner part needs to go a little out and then come up
%		so that the whole assembly is not flimsy and sits perfectly
%   fabric instead of paper so that it can be integrated into clothing
%       direction of fabric, stretchier side in long direction
%   sensor goes over fabric, why we don that

\begin{figure}[ht!]
\centering
\includegraphics[width=1\linewidth]{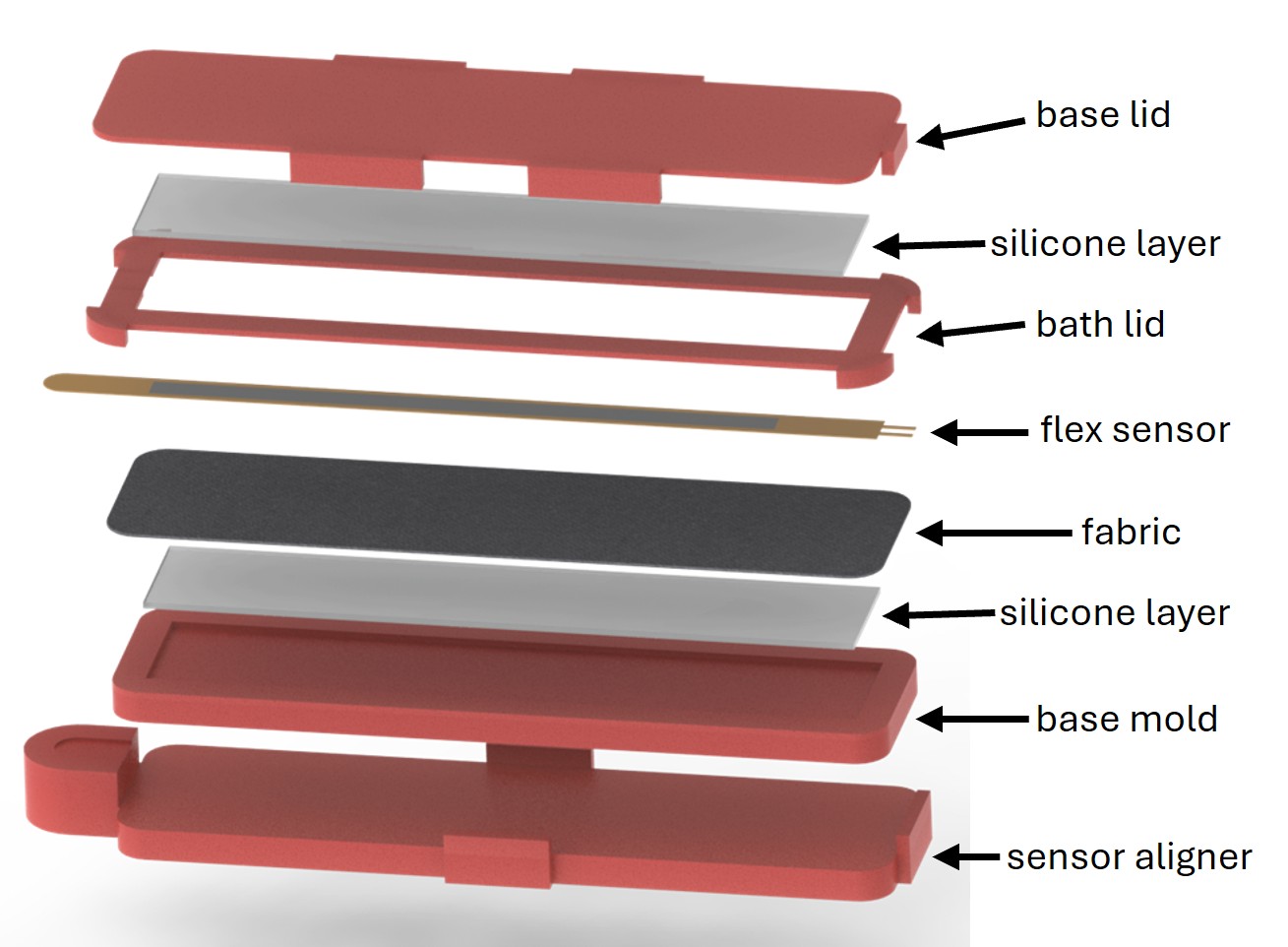}
\caption{Exploded view of the base layer components.}
\label{fig.explodedViewOfBaseLayerStep}
\end{figure}

\begin{figure}[ht!]
\centering
\includegraphics[width=1\linewidth]{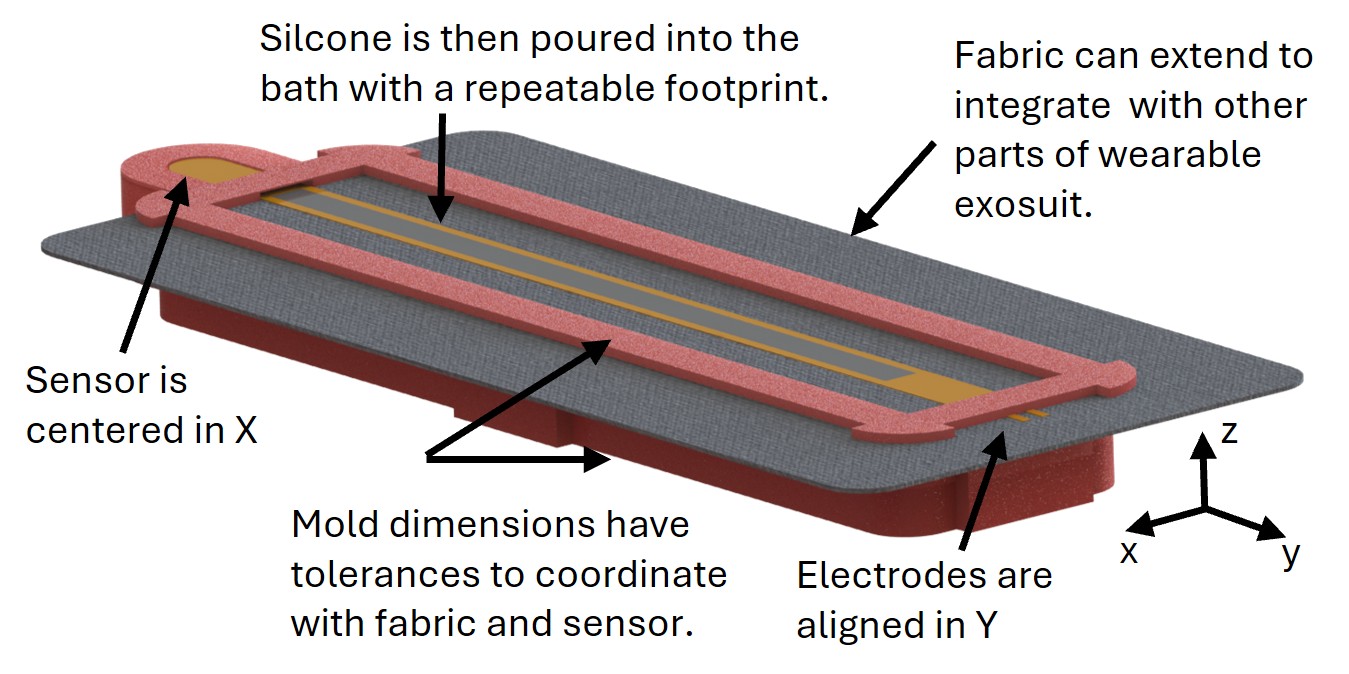}
\caption{Assembled view of the fabric and sensor embedding in the base mold.}
\label{fig.cadExtendedFabric}
\end{figure}

The parts necessary for the fabrication of the base layer are shown in Fig.~\ref{fig.explodedViewOfBaseLayerStep}, which include the sensor aligner, the base mold, the fabric, the flex sensor, the bath lid, and the base lid. We start by cutting a piece of fabric slightly larger than the base mold, as shown in Fig.~\ref{fig.cadExtendedFabric}.
In real-world applications, this fabric might extend to be as large as necessary for integration with a wearable robotic exosuit or other soft robotic form factors.
The fabric also serves as a limiting layer (similar to the traditional use of paper), which allows the fPN to provide rotational (bending) actuation rather than linear (elongation) actuation.
We make sure that the weft dimension of the fabric, which allows more stretch, is oriented along the length of the base mold. The less compliant warp dimension then prevents radial bulging of the actuator's cross-section, which helps stabilize the sensor. The more compliant weft dimension allows a small amount of elongation, which is beneficial for wearable exosuit applications where the fPN must concentrically rotate around a human joint.
%We do this to counteract the actuator's bulging on the sides by embedding the fabric, which acts as a limiting layer.
The base mold is toleranced to have a snap fit with the sensor aligner,
%, which in this case was an 0.3mm clearance in Y to account for PLA expansion of the sensor aligner.
and the bath lid is toleranced to have a snap fit over the base mold and fabric.
%, which in this case was an 0.3mm clearance in X and Y to account for the fabric thickness.
%The bath lid corner lips have an 0.2m depth in Z to secure themselves over the fabric.
The bath lid corner lips have some depth to secure themselves over the fabric.
%A 0.3 mm offset is given in the sensor aligner part to account for the tolerance of 3D printed parts, so that any fit issues don't occur when the base mold and the sensor aligner are assembled, which are designed to have a snap fit.

\begin{figure}[ht!]
\centering
\includegraphics[width=1\linewidth]{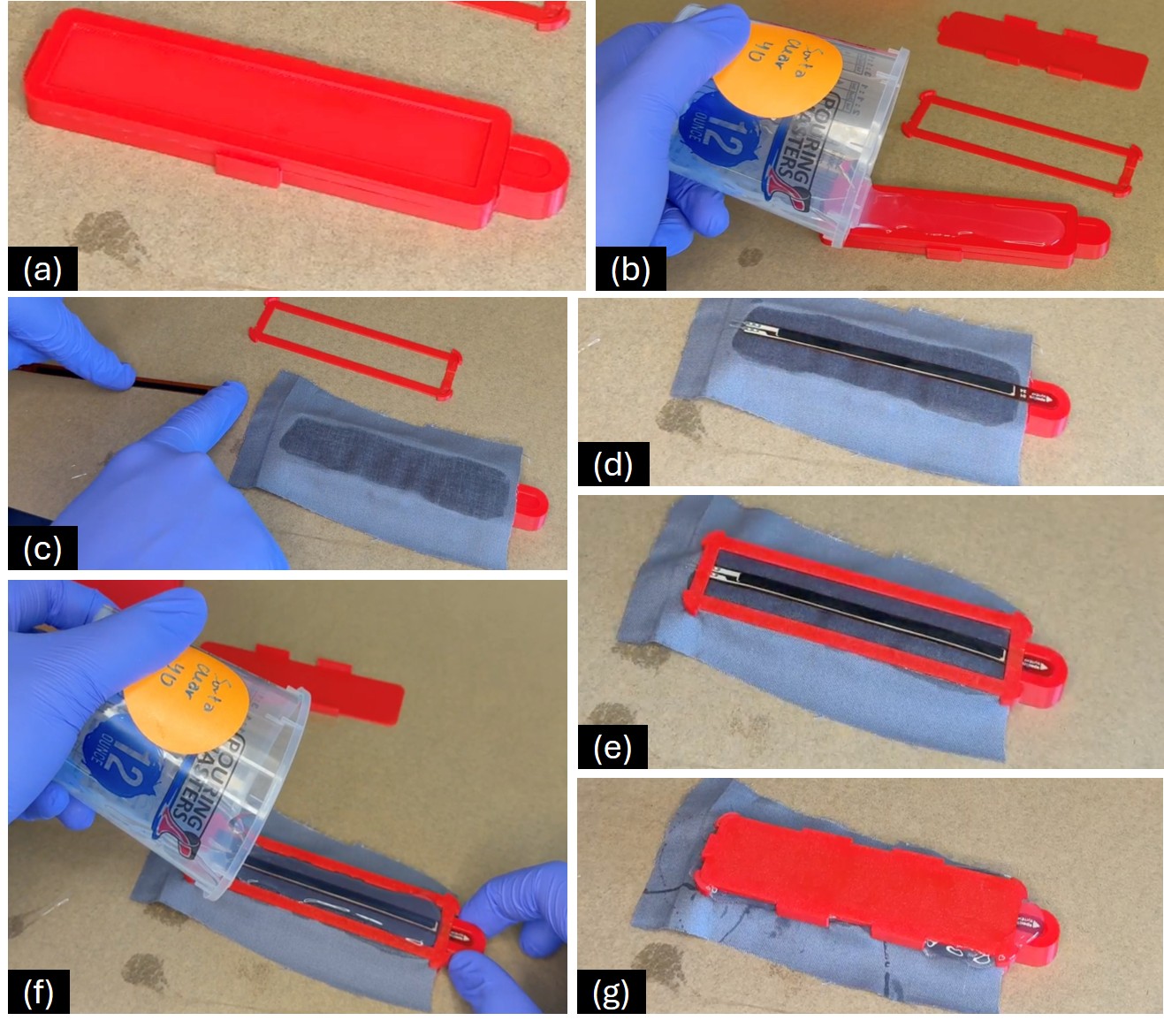}
\caption{Base layer fabrication with sensor embedding.}
%: (a) base layer and sensor aligner assembly, (b) pouring silicone in base layer, (c) placing the fabric on poured silicone and applying silicone on flex sensor, (d) placing the flex sensor on the fabric, (e) placing the bath lid, (f) pouring silicone in the sensor embedding bath, (g) attaching the base lid.}
\label{fig.fabricatingBaseLayerWithSensorEmbedding}
\end{figure}

Fig.~\ref{fig.fabricatingBaseLayerWithSensorEmbedding} shows the steps of the base layer fabrication with sensor embedding.
First, the sensor aligner is assembled with the base mold (Fig.~\ref{fig.fabricatingBaseLayerWithSensorEmbedding}a).
%Once the sensor aligner and the base mold are assembled 
The silicone is then prepared.
SORTA-Clear 40 (Smooth-On, Inc.) is chosen for the soft actuator body due to its transparency, food-grade
quality, high shore hardness, and low-hazard processes. SORTA-Clear 40 is more challenging to seal compared to other Smooth-On platinum cure silicones, making it a good benchmark for the processes presented here. Parts A and B are mixed in a 10:1 ratio as recommended by the manufacturer, and then degassed for 4 minutes in a vacuum chamber to remove air bubbles. 
The degassed silicone is poured into the base mold until the bath of the mold is entirely filled (Fig.~\ref{fig.fabricatingBaseLayerWithSensorEmbedding}b). Once the uncured silicone settles with gravity in the base mold, the fabric is gently placed, and the uncured silicone seeps through the fabric. This seeping process can be sped up by gently stretching the corners of the fabric.

While waiting for the silicone to seep into the fabric, a thin layer of silicone is applied on one side of the flex sensor (Fig.~\ref{fig.fabricatingBaseLayerWithSensorEmbedding}c). The sensor aligner is used to ensure the flex sensor is centered in the X direction and positioned in the  Y direction for the contact leads to have repeatable integration with external circuitry (Fig.~\ref{fig.fabricatingBaseLayerWithSensorEmbedding}d).
The sensor is placed above the fabric instead of beneath it to provide protection of the sensor from the exterior as well as protection of the thin silicone base from tearing due to the sensor's edge.
%to provide a cushion layer because the sensor is not hyperelastic, and if it is below the fabric, there is much less silicone below it without any strain-limiting layer, thus the chances of the actuator failing from the base increase drastically.
The bath lid is fixed onto the assembly (Fig.~\ref{fig.fabricatingBaseLayerWithSensorEmbedding}e). The bath lid has a 1mm depth, which creates a bath to pour a second layer of silicone into (Fig.~\ref{fig.fabricatingBaseLayerWithSensorEmbedding}f), ensuring that the sensor and fabric are embedded completely.
Having the fabric embedded ensures the base will not delaminate from the main body, and having the sensor embedded ensures the sensor will not be exposed to air flow in the internal cavity during pneumatic actuation.
The open window in the bath lid 
%combination of the base mold and the bath lid
creates a repeatable embedding footprint.
Since the fabric acts a functional strain-limiting layer affecting bending behavior during actuation, this allows for repeatable functional performance.
Together with the sensor aligner, this footprint promotes repeatable readings from the sensor and prevents the sensor contact leads from becoming embedded.
The base lid is then placed and pressed gently to ensure excess silicone rising above the 1mm bath is flushed out (Fig.~\ref{fig.fabricatingBaseLayerWithSensorEmbedding}g). This allows for the mold to cure with a flat surface, which is crucial for avoiding air leaks in the sealing layer, as discussed in the next step. The silicone is cured overnight and then partially demolded by removing the base lid.

\subsection{Fabrication of Main Body with Rounded top and External Cuts}\label{subsec.main}
%fabrication of main body
%	other things
%		demold parts
%		fabrication of initial tubing
%air inlet
%  made of dragon skin because it seals very well to other silicones
%    not as rigid, ziptie
%    ecoflex is too soft, you get bulging after the ziptie

\begin{figure}[ht!]
\centering
\includegraphics[width=.9\linewidth]{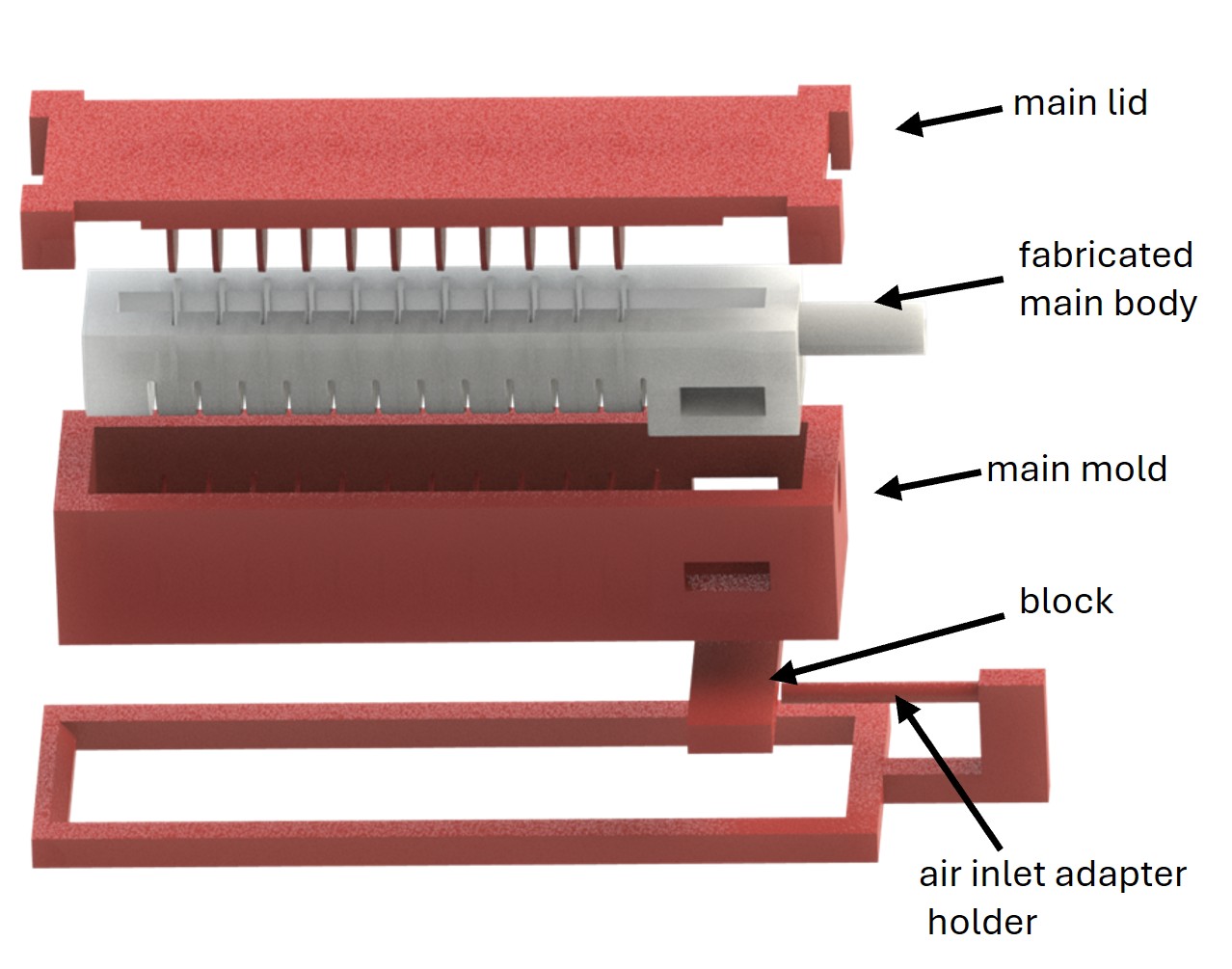}
\caption{Exploded view of the main body components.}
\label{fig.mainCADExploded}
\end{figure}

The main body is cast in parallel with the base. Fig.~\ref{fig.mainCADExploded} shows the parts required in this step, which include the air inlet adapter holder, the block, the main mold, and the main lid.
The block results in an external cut in the cast silicone (labeled in Fig.~\ref{fig.methodsIntro}) for assembly in a larger system, such as clamping to a test rig for running experiments (see Section~\ref{sec.results} Fig.~\ref{fig.expActuator}), or for rigid integration into a robotic system. The main body is cast with the bottom facing up. The surface of the rounded top is 3D printed into the rounded floor of the main mold. In turn, the mold features for the external cuts are 3D printed into the main mold, and the mold features for the internal air chambers are 3D printed into the main lid.

\begin{figure}[ht!]
\centering
\includegraphics[width=1\linewidth]{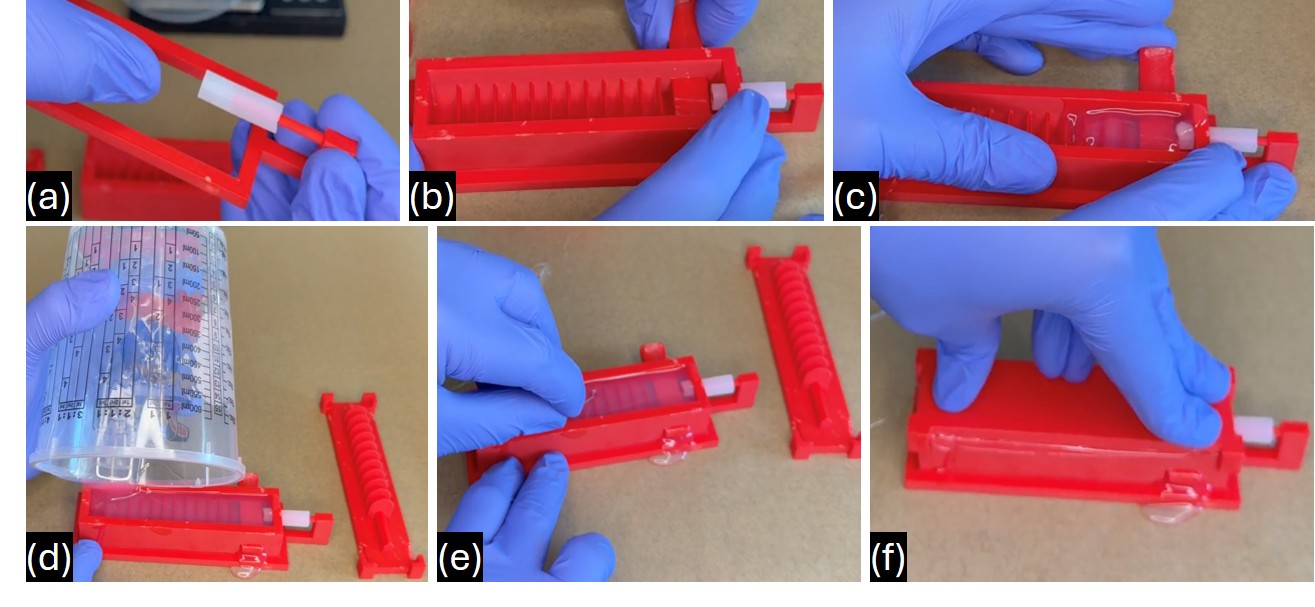}
\caption{Steps for the main body fabrication.}
%: (a) attaching the air inlet adapter, (b) inserting the block in the main mold, (c) pouring silicone underneath the block and threading the block through, (d) finishing the pour of the silicone in the main mold, (e) fishing out air bubbles, (f) closing main lid.}
\label{fig.fabricatingMainBody}
\end{figure}

The detailed fabrication process is shown in Fig.~\ref{fig.fabricatingMainBody}. First, the pre-fabricated air inlet adapter is attached to the air inlet adapter holder (Fig.~\ref{fig.fabricatingMainBody}a), and this subassembly is then attached to the main mold. The block is partially threaded through the opening in the main body mold (Fig.~\ref{fig.fabricatingMainBody}b).
SORTA-Clear 40 silicone is mixed and degassed as described in Section~\ref{subsec.base}, and then poured into the main body mold.
Once the silicone rises above the block (Fig.~\ref{fig.fabricatingMainBody}c), the block is pushed entirely through the main body.
The mold is filled with the remaining silicone (Fig.~\ref{fig.fabricatingMainBody}d), and any air bubbles are removed with the end of a paper clip wire (Fig.~\ref{fig.fabricatingMainBody}e). The main lid is placed and gently pressed so that extra silicone flows out through the sides (Fig.~\ref{fig.fabricatingMainBody}f). The pressing is gentle to minimize disturbance of the silicone in the main body mold and ensure no air bubbles are introduced. The assembled mold is cured overnight and then the fabricated part is demolded. A thin film of silicone will form at the air inlet adapter opening, which is removed using blunt plastic tweezers to avoid damaging the cured silicone and ensure a proper opening for the pneumatic tube to be inserted.

\subsection{Fabrication of the Sealing Layer}\label{subsec.sealing}
%robust sealing layer
%	creating small sealing bath of the same size, getting rid of uneveness
%	getting rid of human error
%	make sure not to apply pressure on terminals of sensor
%	other things
%		demold
%		insert tubing

The presented fabrication process of the sealing layer is crucial for ensuring a robust air-tight seal as well as preventing clogging of the internal air cavities.
Air leaks can occur if the sealing layer does not fully bond the top and base.
Clogging of the internal cavities can occur due to capillary action of the sealing layer's uncured silicone rising up along the inner walls of the cured main (see Fig.~\ref{fig.cloggingRealWorld}).
%This is the most crucial part of the whole process. If the sealing is not done correctly, it will result in air leaks or silicone clogging due to capillary action. We show here the new sealing technique that gives a robust and repeatable process to produce airtight fast pneumatic actuators.

\begin{figure}[ht!]
\centering
\includegraphics[width=.9\linewidth]{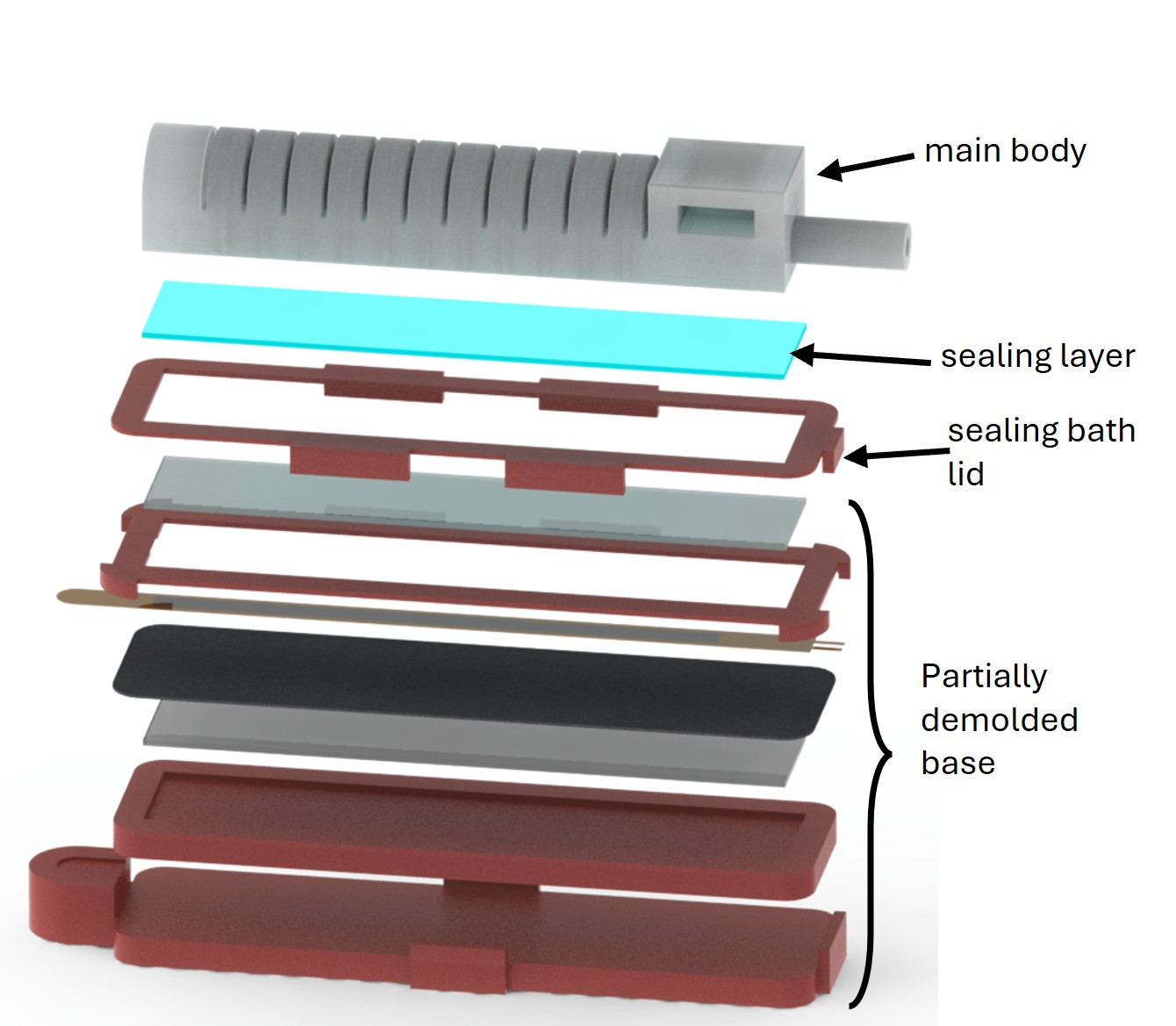}
\caption{Exploded view of the sealing layer components.}
\label{fig.sealingLayerExplodedCAD}
\end{figure}

\begin{figure}[ht!]
\centering
\includegraphics[width=.9\linewidth]{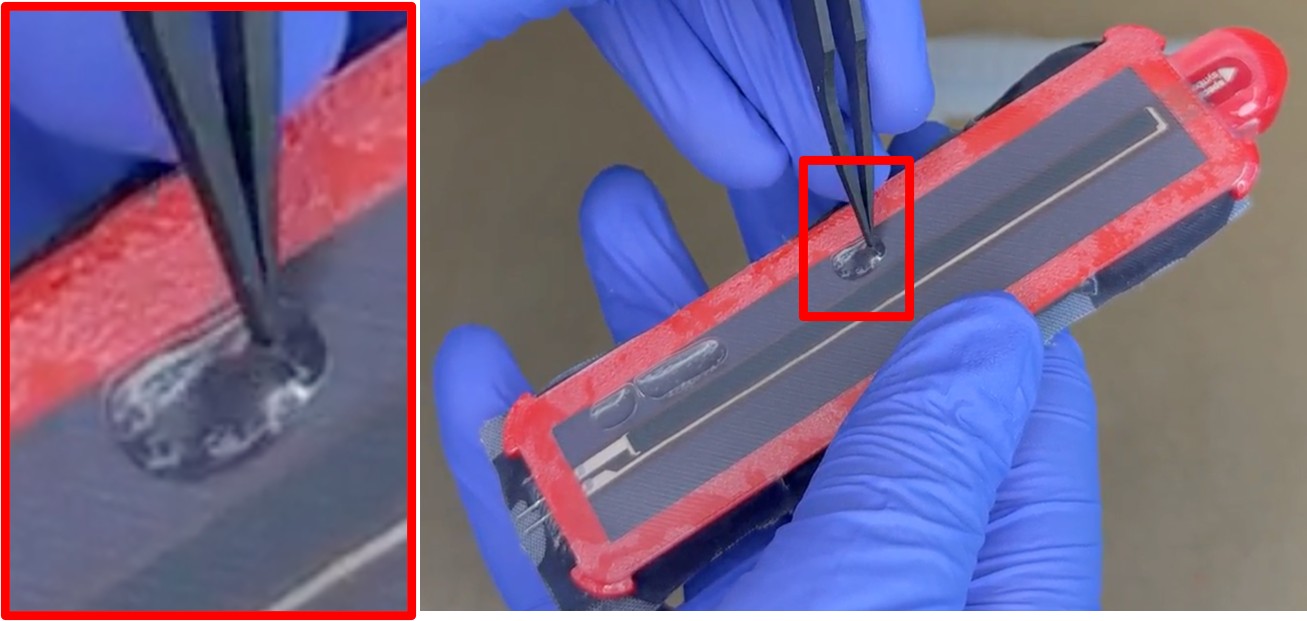}
\caption{Exposing air pockets in the base layer after curing.}
\label{fig.removingAirBubblesBeforeSealing}
\end{figure}

The parts used in the sealing layer fabrication are shown in the exploded view in Fig.~\ref{fig.sealingLayerExplodedCAD}. 
This includes the cured base, which has been partially demolded by removing the base lid, as well as the sealing bath lid and the cured demolded main body.
The first step is to remove the thin silicone films covering air pockets which may have formed on the base during curing, as shown in Fig.~\ref{fig.removingAirBubblesBeforeSealing}. This is done using blunt plastic tweezers.
The air pockets form when placing the base lid on top of the uncured silicone.
The uncured silicone sticks to the PLA of the base lid, causing air pockets when air becomes entrapped. 
Although not ideal, the air pockets are easily filled with the next layer of silicone. 
The base lid is necessary to prevent the meniscus of the highly viscous silicone.
A flat surface of silicone is necessary for controlling the volume of the sealing layer in the next step.

%\textcolor{red}{When placing the base lid, the uncured silicone in the bath sticks to the PLA of the base lid, which causes air pockets to force since the air does not have room to escape. However,} these shallow air pockets will be easily filled with the sealing-layer silicone.

\begin{figure}[ht!]
\centering
\includegraphics[width=1\linewidth]{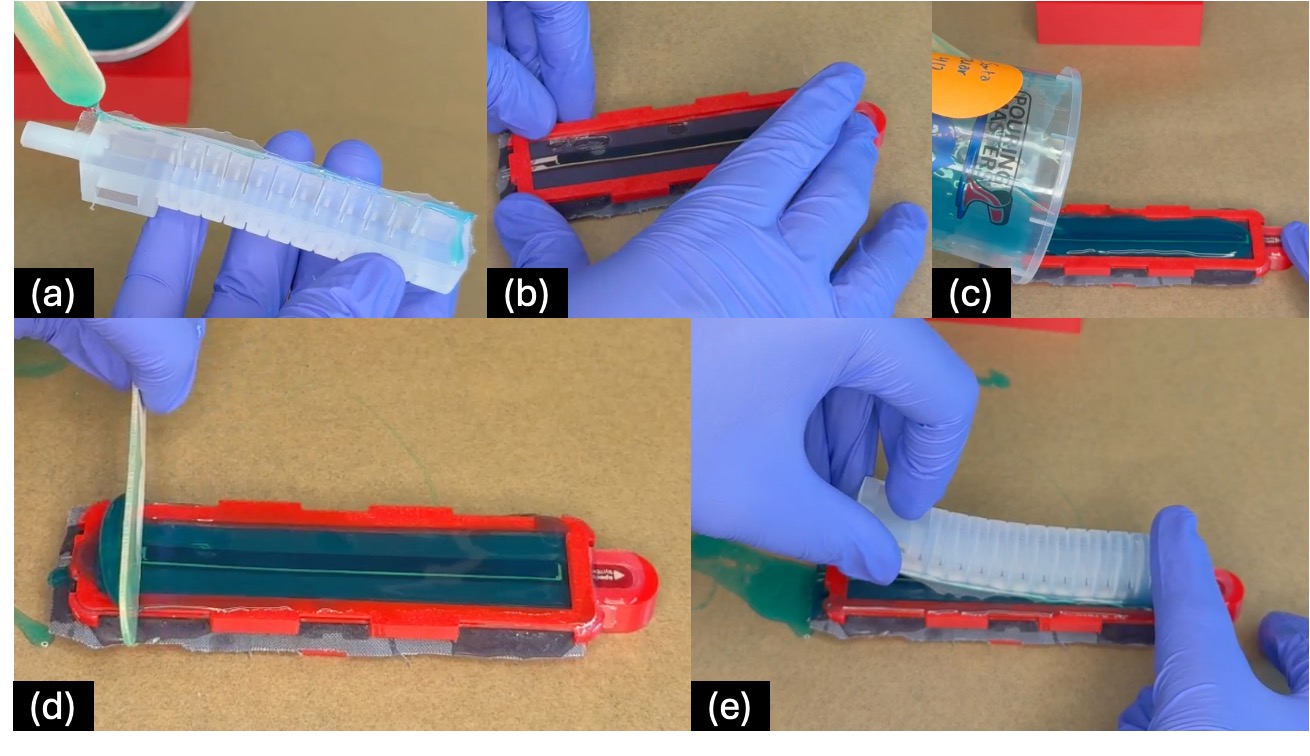}
\caption{Steps for the sealing layer fabrication: (a) applying a thin layer of silicone to the main body, (b) placing the sealing bath lid on the base mold assembly, (c) pouring silicone in the sealing bath, (d) removing excess silicone in the sealing layer bath, (e) placing the main body in the sealing bath.}
\label{fig.sealingLayerSteps}
\end{figure}

\begin{figure}[ht!]
\centering
\includegraphics[width=.9\linewidth]{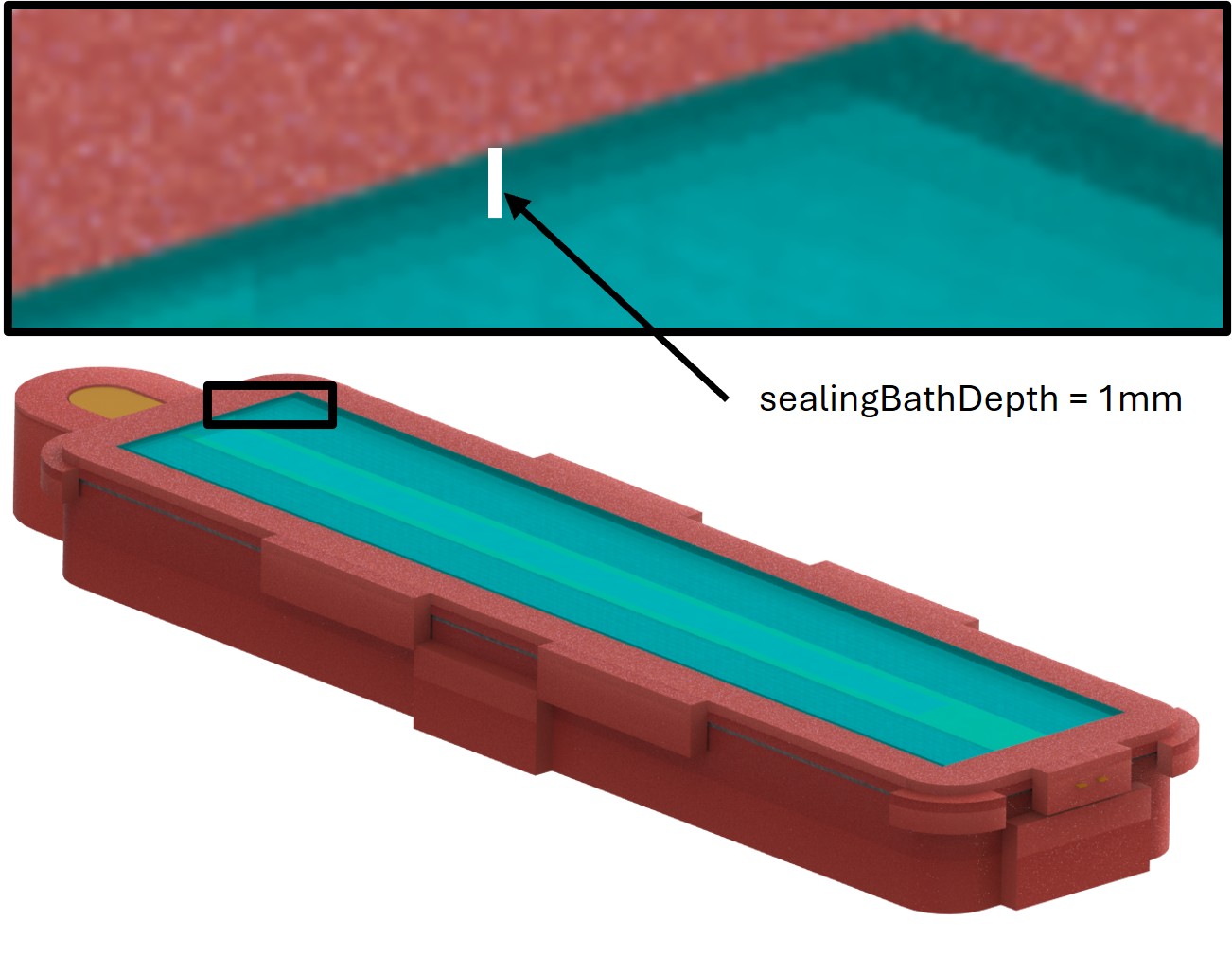}
\caption{The controlled sealing bath.}
\label{fig.sealinglayerCAD}
\end{figure}

The fabrication steps for the sealing layer are shown in Fig.~\ref {fig.sealingLayerSteps}. 
SORTA-Clear 40 silicone is mixed and degassed as described in Section~\ref{subsec.base}.
Teal-colored pigment (Silc Pig, Smooth-On, Inc) is added to the uncured silicone before mixing, to help track unwanted fabrication errors.
This pigment was crucial in developing the procedures presented here, to empirically determine the cause of fabrication defects.
Additionally, the pigment allows for automated color thresholding for experimental testing (see Section~\ref{subsec.calibration}).
A thin layer of the uncured silicone is applied to the cured main body, on the surface that exposes the air cavity (Fig.~\ref {fig.sealingLayerSteps}a). This is done to aid the sealing process. The sealing bath lid is attached to the base layer assembly (Fig.~\ref {fig.sealingLayerSteps}b). To enable repeatability of the sealing process, the dimensions of the sealing bath and the fPN actuator are matched.
The uncured silicone is poured into the bath as done for the base mold (Fig.~\ref {fig.sealingLayerSteps}c). Once the silicone fills the sealing bath and settles into the corners, a craft stick is slid over the top surface of the sealing bath mold to get rid of extra silicone above the bath (Fig.~\ref {fig.sealingLayerSteps}d). The sealing bath is 1~mm deep, as shown in Fig.~\ref{fig.sealinglayerCAD}.
%This is similar to the bath lid in the base mold, which was also 1mm deep.
Controlling the depth of the sealing layer to exactly 1 mm allows for a deep enough bath to avoid sealing failure and air leaks, while still being shallow enough to avoid clogging due to capillary action.
Furthermore, the controlled uniform depth minimizes the chance of deep or shallow non-uniform patches caused by human error. The 1mm depth dimension was empirically determined through fabrication experiments.
Note that not only does the sealing bath lid enable this controlled bath height, but it also allows the process to be robust against human error caused by changes in speed and pressure of sliding the craft stick.

%allowing for a robust sealing with just enough silicone so that capillary action does not occur. This sealing technique allows for the same level of silicone at each point, thus getting rid of any human-induced error when ensuring the main body comes into contact with an equal amount of silicone in the bath.
The main body is then placed in the sealing bath (Fig.~\ref {fig.sealingLayerSteps}e). It is placed gently to avoid disturbance, as any unnecessary movement in the uncured silicone can cause air bubbles to form.
%As an additional safety measure, silicone is applied on the mating edges of the main body to ensure better sealing (Fig.~\ref {fig.sealingLayerSteps}f). 
The uncured sealing layer in the bath will slightly rise up above the bath walls, helping to encase the main body for a robust seal. The amount of raised sealing height will be controlled and repeatable by the initial volume of the bath.
The assembly is cured overnight at room temperature and then demolded. Room temperature curing ensures the air inside the internal soft actuator cavity does not expand and compromise the seal. Scissors are used to trim excess material. A pneumatic tube is inserted and secured in the air inlet adapter for pneumatic actuation.
%Save this for a future paper.
%Since the air inlet adapter is made of dragon skin, a softer silicone, it allows for better sealing than SORTA-Clear 40. Also, using Dragon Skin over Eco Flex helped resolve the adapter's bulging issue.

%------------------------------------------------------------------
\section{Experiments and Results}\label{sec.results}
%------------------------------------------------------------------
%FEM
%	bending angle
%		vertical rig
%	force
%		horizontal rig
%exp results
%	bending angle
%		(mostafa) angle vs pressure
%			computer vision to show angle
%			show images we didn't have room for 
%	force
%		horizontal rig

\subsection{Finite Element Analysis of Stress and Bending Behavior}
\begin{figure}[ht!]
\centering
\includegraphics[width=.9\linewidth]{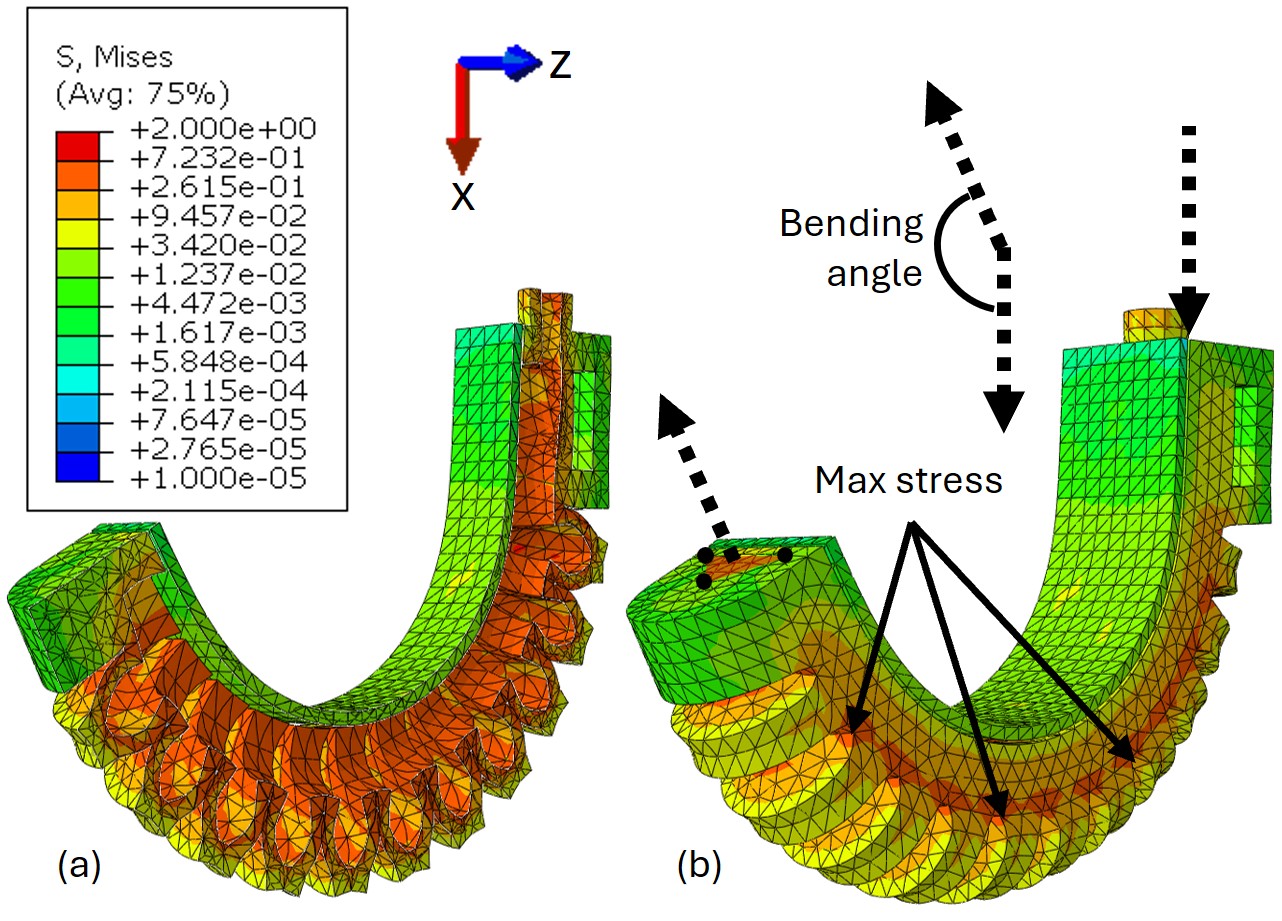}
\caption{Abaqus FEM Simulation. (a) Sectional view. (b) Full view with bending angle and max stress depicted.}
\label{fig.resultsFEM}
\end{figure}

A Finite Element Model (FEM) of the fPN actuator was performed with Abaqus Standard (Dassault Syst\`emes). The SORTA-Clear 40 material is modeled as a 3rd order Yeoh model, with coefficients C1 = 0.1~MPa, C2 = 0.211~MPa, C3 = 0.00166~MPa, and density =~\SI{1.08e-9}{\tonne\per\mm\cubed}, derived from ~\cite{marechal2021toward}. The actuator is modeled as an assembly of three layers: the main body, the base layer, and the strain-limiting layer in between.  The main and base layers are modeled with SORTA-Clear 40. The strain-limiting layer is modeled as linearly elastic with density =~\SI{7.5e-10}{\tonne\per\mm\cubed}, Young's modulus = 6500 MPa, and Poisson's ratio = 0.2 derived from~\cite{Holland2014SoftRoboticsToolkit}.
%is sandwiched between a base layer and main body made of SORTA-Clear 40.
%The pressure in the air cavity is applied equally on the inner surfaces as a ramp input with a maximum pressure of 20 psi.
%Two boundary conditions are used: 1. 
The actuator is fixed at the air inlet side using an encastre boundary condition.
%, and 2. The actuator is only allowed to rotate about the z-axis and not translate along the z-axis.
%This second boundary condition accounts for asymmetry due to numerical approximations in the mesh, which can cause unwanted twisting deformation.
%at the centroidal element positions.
A uniform pressure is applied to the internal walls of the air cavity. The pressure is applied as a ramp input form 0 to 20 psi (0.138 MPa).
The true deformation at 20 psi is shown in Fig.~\ref{fig.resultsFEM}.
The equivalent stress and the centroids of the nodal coordinates are tracked over the ramp from 0 to 20 psi.
%Fig.~\ref{fig.resultsFEM} shows the final results from the FEM simulation, which gives an estimate of the behavior.
%The maximum Von Mises stress on the internal silicone walls is observed to be 1.588 MPa, whereas the overall maximum Von Mises stress of 1.664 MPa is on the external surface of the actuator.
As shown in Fig.~\ref{fig.resultsFEM}b, the maximum equivalent stress is on the external surface of the actuator walls, concentrated at the bottom corners of the external chamber cuts.  
This is because the highest deformation occurs here due to the geometry.
%This observed difference in the internal and external parts of the fPN actuator is the result of the bending that occurs in the hyperelastic material, creating more deformation externally rather than internally.

\begin{figure}[ht!]
\centering
\includegraphics[width=.9\linewidth]{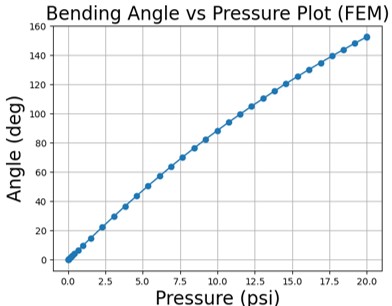}
\caption{Predicted bending angle vs internal Pressure from FEM. Bending angle is post-processed from nodal outputs.}
\label{fig.angleVsPressurePlot}
\end{figure}

Three nodes are tracked on the free end of the actuator (Fig.~\ref{fig.resultsFEM}b) at the assembly level to measure the bending angle.
%MATLAB is used to post-process the output logs.
The surface normal of the free end is calculated from the three nodes, and a bending angle is calculated from the surface normals of the fixed and free ends.
%Once the Abaqus simulation is done, for post-processing the data, the Abaqus2Matlab plugin is used ~\cite{papazafeiropoulos2017abaqus2matlab}, where the .odb file is used as an input.
%The three tracked nodes are used to create a plane, and the angle is derived between the normal of this plane and the fixed end plane, which gives us the bending angle.
The predicted bending angle in simulation reaches 153$^\circ$ at 20 psi, as shown in Figure~\ref{fig.angleVsPressurePlot}.
This follows real-world experiments to a large extent (see Section~\ref{subsec.control}), deviating somewhat at higher pressures. %This may be attributed to not accounting for the embedded flex sensor and anisotropy of the fabric.
%This follows the same trajectory as the real-world experiments but deviates from the observed bending angle from the flex sensor and the computer vision model at higher pressures. Reducing the element size from 5.4 to 2.5 shows improvement in the similarity to real world, however, it requires extra compute power as we reduce the element size. Our next steps would be to improve the simulation results by tweaking the element size to close the sim-to-real gap.

\subsection{Flex Sensor Calibration Using Image-Based Angle Measurements}\label{subsec.calibration}
\begin{figure}[ht!]
\centering
\includegraphics[width=.9\linewidth]{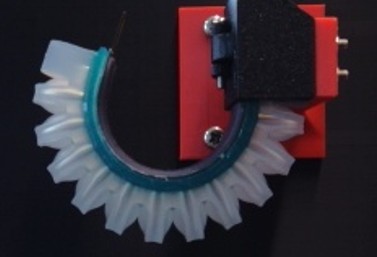}
\caption{One frame from a sequence of images used for angle estimation to calibrate the flex sensor. The experimental setup depicts the fPN being pneumatically actuated with PID pressure control to a reference of 20~psi.}
\label{fig.expActuator}
\end{figure}

To calibrate the embedded flex sensor, the actuator bending angle was measured using an image-based approach. The actuator was mounted vertically against a high-contrast background as shown in Fig.~\ref{fig.expActuator}, and pneumatic pressure was increased from 0 to 20 psi in 0.5 psi steps. At each pressure level, the images were captured.
%as the actuator reached the transient and steady-state stages.

\begin{figure}[ht!]
\centering
\includegraphics[width=.9\linewidth]{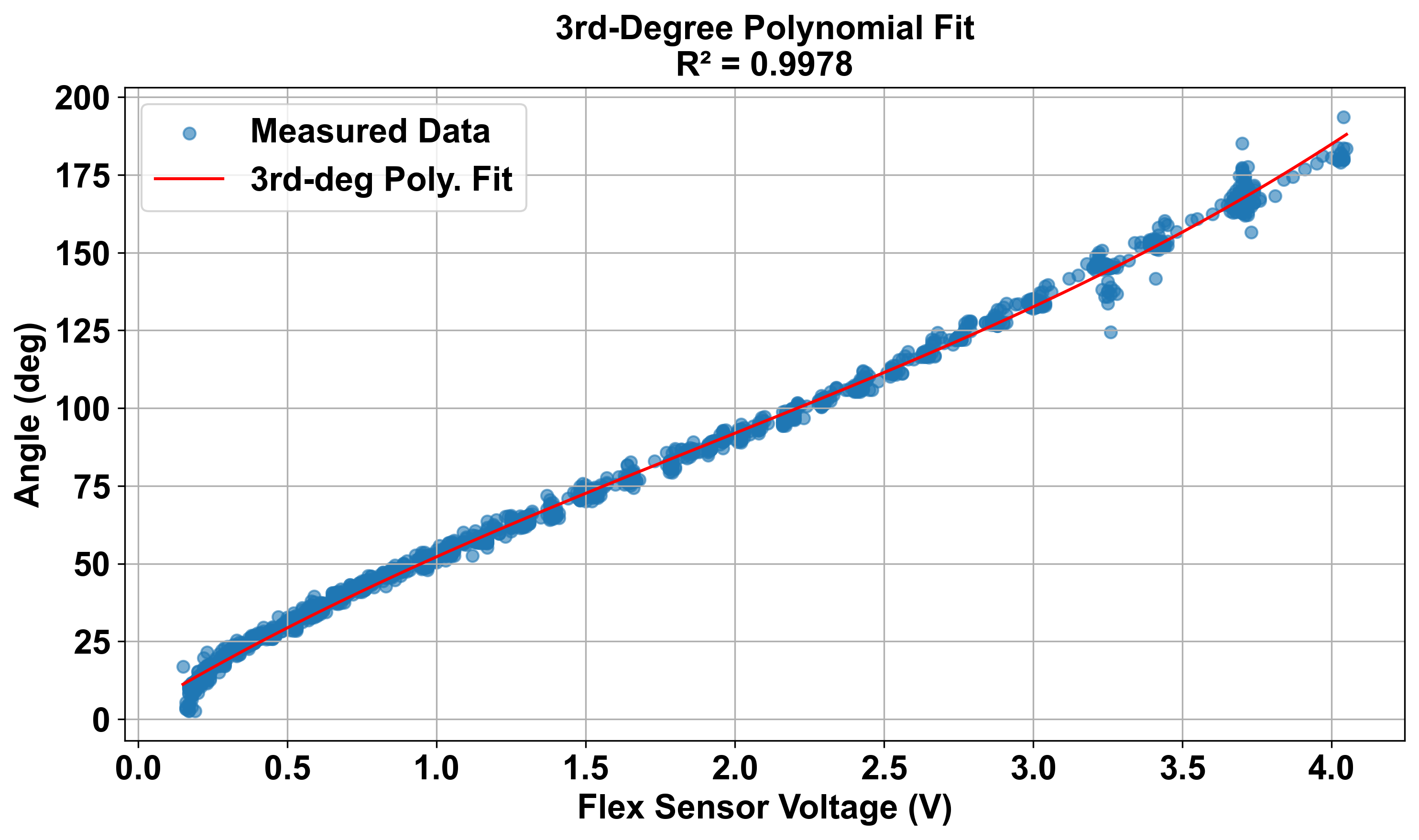}
\caption{Flex sensor calibration using image-based tip angle measurement.}
\label{fig.flexSensorCalb}
\end{figure}

Using MATLAB’s image processing toolbox, the teal-colored sealing layer was extracted from each image via thresholding and edge detection. The tip orientation was then computed relative to the fixed base, yielding the corresponding bending angle for each pressure state. This process resulted in a dataset relating sensor voltage to actuator bending angle over the full operating range, shown in Fig.~\ref{fig.flexSensorCalb}.

A 3\textsuperscript{rd}-order polynomial was fit to the voltage–angle data, resulting in the following calibration equation:

\begin{equation}
\theta = 1.7874 V^3 - 10.2474 V^2 + 57.9351 V + 2.7223
\end{equation}

where $\theta$ is the actuator bending angle in degrees and $V$ is the flex sensor voltage in volts. The fit showed excellent agreement with the experimental data, achieving an $R^2$ value of 0.9978, RMSE value of 2.3054, and was used for real-time angle estimation in subsequent experiments.

\subsection{Pressure Control Performance and Angle Response}\label{subsec.control}
A closed-loop PID controller was implemented to regulate the internal pressure of each soft pneumatic actuator. The pneumatic control station published in~\cite{Massoud2024-IEEEAccess, Massoud2025-RAL} was used for pressure regulation; however, the original 3/2 on/off valve configuration was replaced with two 2/2 proportional valves to reduce pressure fluctuations and improve control smoothness. Accurate pressure control is necessary to ensure consistent bending behavior. Feedback is provided by a pressure sensor, and the controller modulates the pneumatic input to track a desired pressure reference.

\begin{figure}[ht!]
\centering
\subfloat[]{\includegraphics[width=0.48\linewidth]{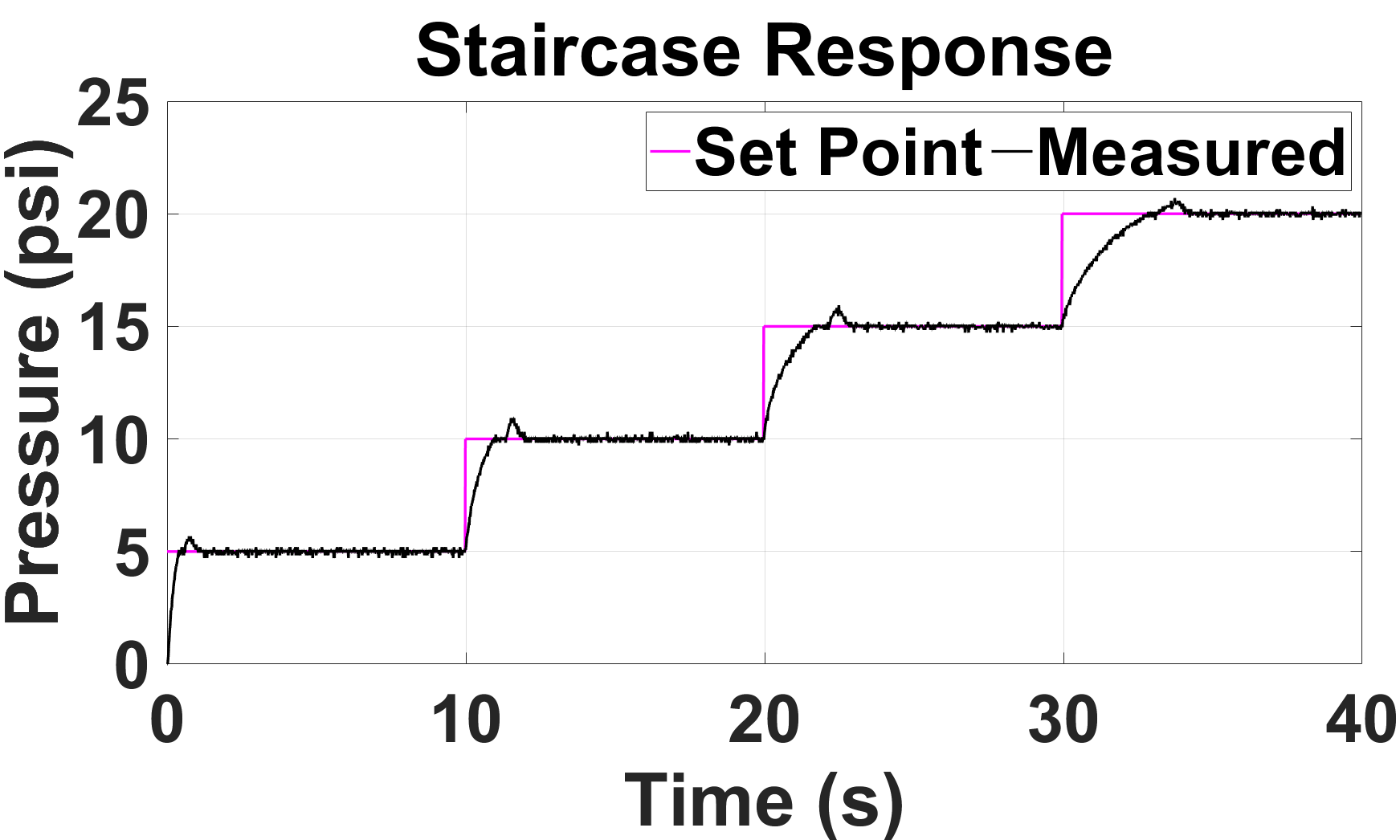}
\label{subfig.stairCase}}
\hfil
%\hspace{-0.4cm}
\subfloat[]{\includegraphics[width=0.48\linewidth]{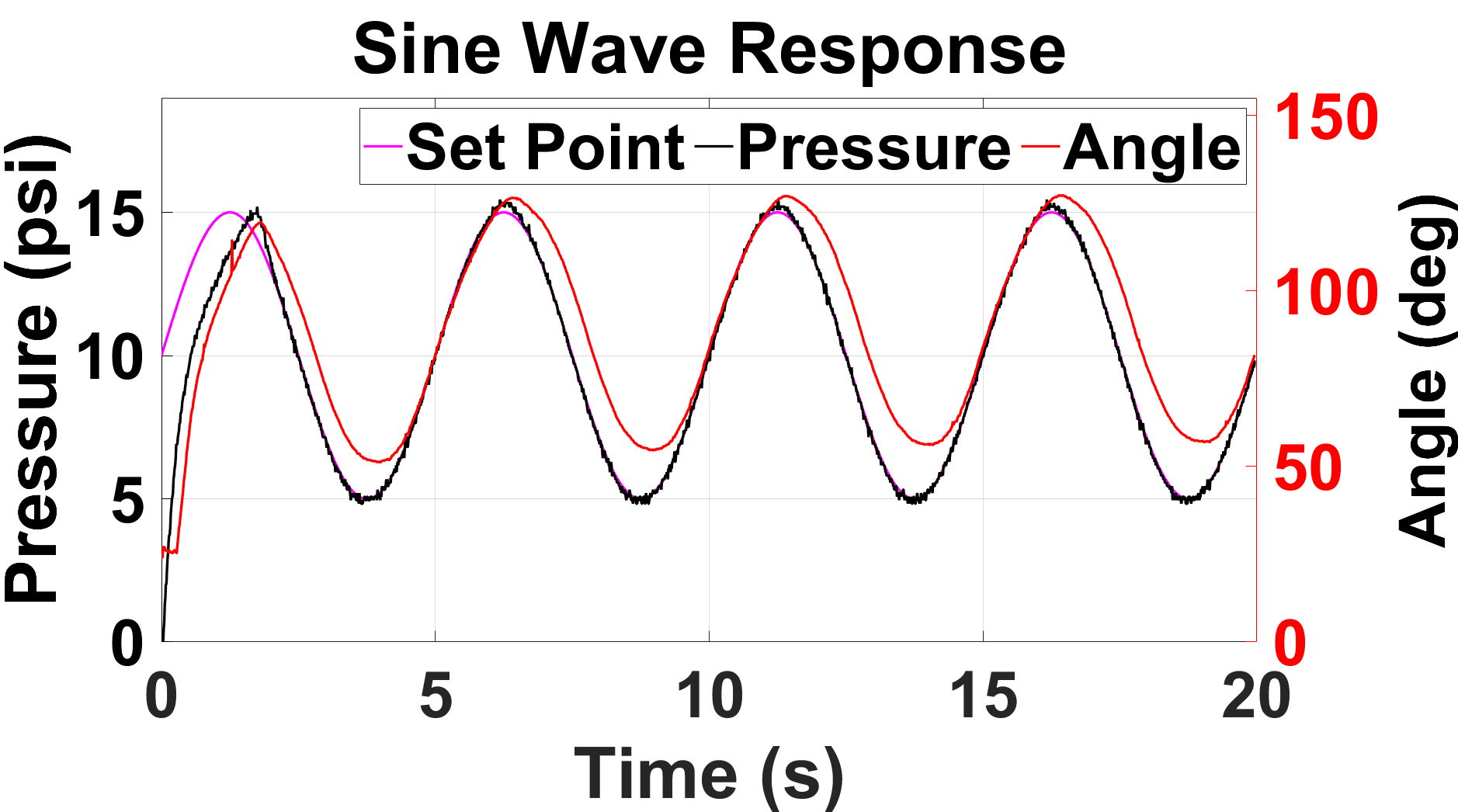}
\label{subfig.sineWave}}
\caption{PID pressure control performance. (a) Staircase pressure tracking response with increasing pressure steps. (b) Sinusoidal pressure tracking response.}
\label{fig.PIDperformace}
\end{figure}

The performance of the PID-based pressure controller was evaluated using both staircase and sinusoidal pressure inputs. As shown in Fig.~\ref{subfig.stairCase} and \ref{subfig.sineWave}, the controller accurately tracks the desired pressure profiles with a rise time of 0.2853~s,  a settling time of 0.8773~s, an overshoot of 13.6~\%, a peak time of 0.7388~s, and a steady-state error of 0.0112~psi. The staircase input demonstrates stable convergence at each pressure level, while the sinusoidal input highlights the controller’s ability to follow dynamic pressure commands.

\begin{figure}[ht!]
\centering
\subfloat[]{\includegraphics[width=0.48\linewidth]{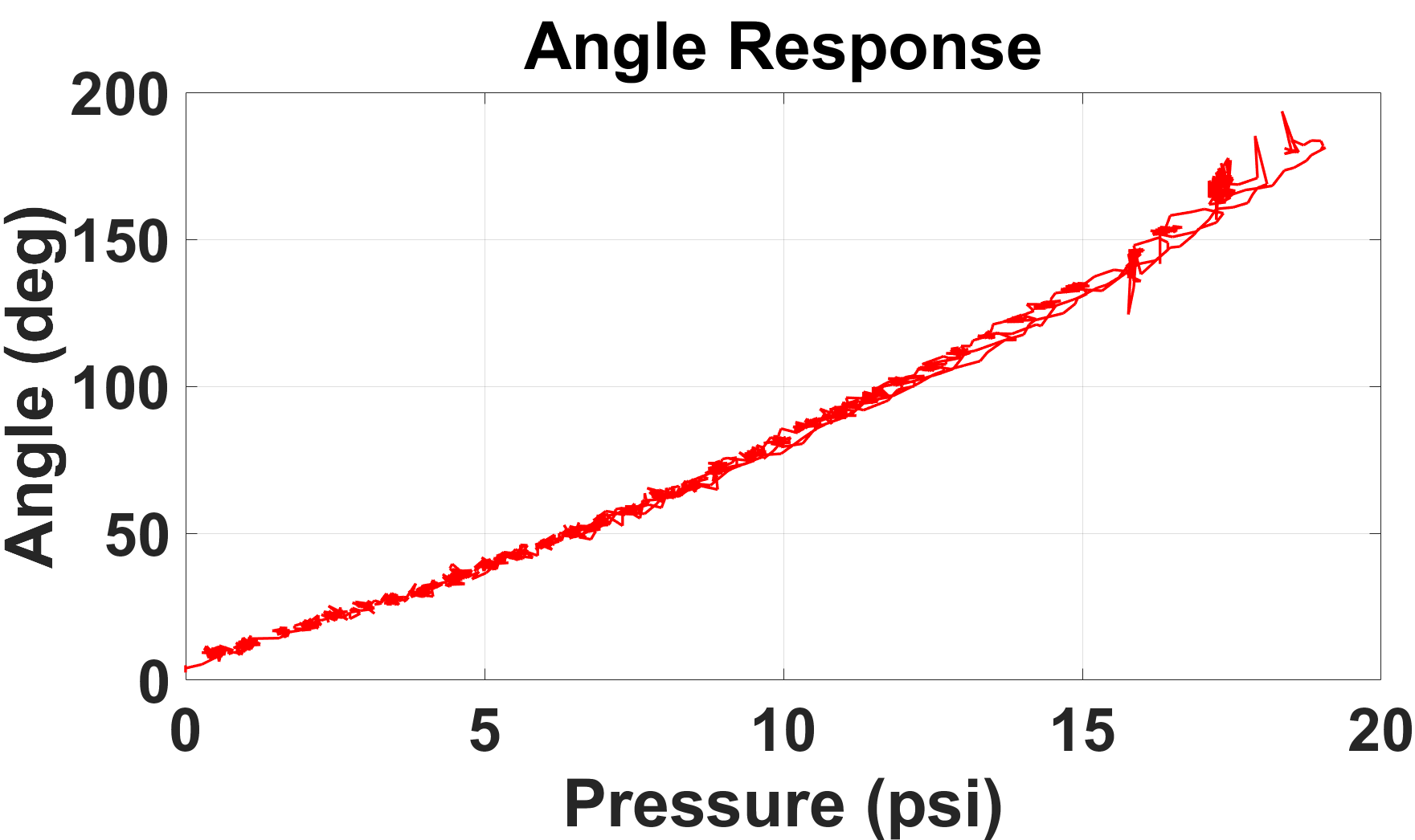}
\label{subfig.stairCaseAngleVsPressure}}
\hfil
\hspace{-0.24cm}
\subfloat[]{\includegraphics[width=0.48\linewidth]{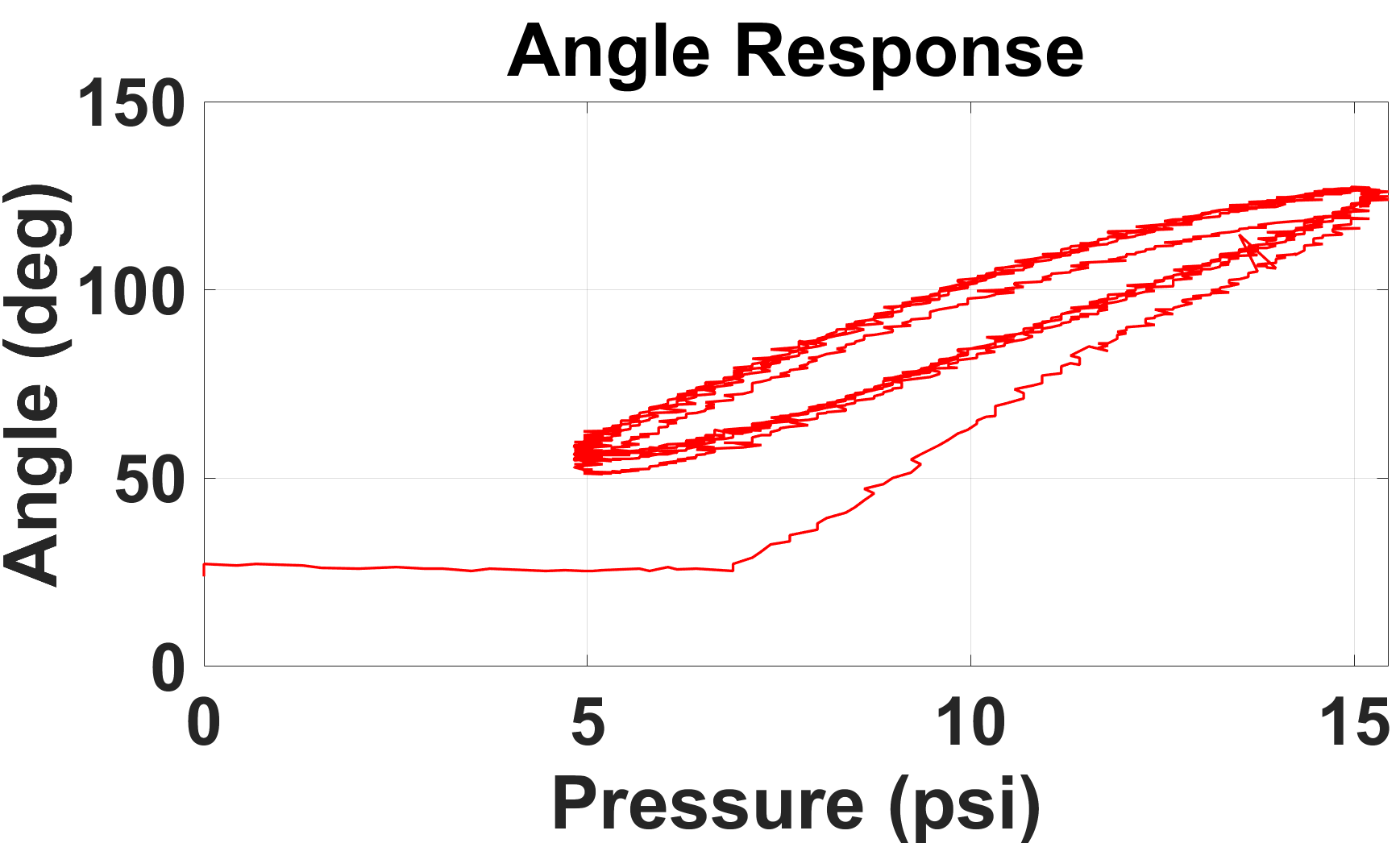}
\label{subfig.sineWaveAngleVsPressure}}
\caption{Actuator bending angle response to pressure input. (a) Angle response under staircase pressure input with 0.5 psi steps. (b) Angle response under sinusoidal pressure input, showing a small amount of hysteresis, with repeatable behavior. Each sine wave cycle is one inflation-deflation loop.}
\label{fig.hysteresis}
\end{figure}

The actuator angle responses are shown in Figs.~\ref{subfig.sineWave} and \ref{fig.hysteresis}.
%The staircase in Fig.~\ref{subfig.stairCaseAngleVsPressure} was obtained by sweeping the pressure from 0 to 20~psi in 0.5psi increments, and then from 20~psi back down to 0.
%The sine wave response in Fig.~\ref{subfig.sineWaveAngleVsPressure} was obtained with an 5 sec period from 0 
The staircase in Fig.~\ref{subfig.stairCaseAngleVsPressure} was obtained by sweeping the pressure from 0 to 20~psi in 0.5~psi increments.
Under the staircase pressure input, the actuator exhibits a smooth and repeatable pressure–angle relationship, which is comparable to the predicted FEM bending angle response in Fig.~\ref{fig.angleVsPressurePlot}, deviating somewhat at higher pressures.
The sine wave in Fig.~\ref{subfig.sineWaveAngleVsPressure} was obtained with a 5 to 15~psi peak-to-peak pressure with a 5~sec period.
For the sinusoidal pressure input, a minimal amount of hysteresis is observed compared with prior experiments with slow pneumatic networks (sPNs)~\cite{Libby2023-arXiv}.
This hysteresis reflects the viscoelastic behavior of the soft actuator and pneumatic dynamics.
Fig.~\ref{subfig.sineWaveAngleVsPressure} nicely demonstrates the repeatable behavior of the actuator under constant loading and unloading cycles.

%/************** For Prof. Libby ********************
%\begin{figure}[ht!]
%\centering
%\includegraphics[width=1\linewidth]%{figuresPNG/sineWaveAngleVsPressureForProfLibby.png}
%\caption{insert Caption.}
%\label{fig.hystBegOfDay}
%\end{figure}

%----------------------------------------------------------------
\section{CONCLUSIONS}
%----------------------------------------------------------------
%First par
%   Review the main points.
%   State contributions again.
%   Similar to last par of intro, but with more technical detail
%       because now they have read the paper.
%   Do not copy the abstract, but rather elaborate.

%2nd par: limitations 
%   In general, you don't mention every new future idea you have.
%   You mention only the ones that are directly related to solving 
%       your limitations, so that the reviewer knows that you are aware.
%       Then you say how you would solve those limitations in the future.
%	Air bubbles in base, maybe due to the lid.
%		Lid is necessary to create a flat surface
%       with the fabric embedded
%		while keeping the base as thin as possible.
%   While the air bubbles are remedied with the sealing layer and hence do
%       not reduce robustness or functionality, we will still
%       investigate methods in the future to remove them
%       from the process.
%   Manually pushing the silicone around the edges of the top layer %    %      (Fig.~\ref{fig.sealingLayerSteps}f)
%       This manual step could introduce air bubbles and non-repeatability.
%       One solution could be increasing sealing bath depth, but a deepr bath could cause clogging.
%   The bath depth should therefore be optimized to handle this tradeoff, and this needs to be done for different siliconed mixed viscosities and internal cavity geometries.

%3rd par: Suggest applications and extensions.
This paper introduces
a fabrication pipeline
%mold parts needed
for achieving a repeatable two-part pour casting of soft pneumatic actuators with robust sealing layers and sensor embedding of commercially available thin-film flex sensors. 
The sensor embedding ensures repeatability and eliminates noise and human error.
The sealing process ensures a controlled and uniform thickness, which minimizes damage, air leaks and clogging.

One shortcoming in the current sealing layer process is that
%\textcolor{blue}{silicone is manually applied on the edges of the main body for a better seal, which can generate air bubbles at that sealing surface and is prone to human error.
%This can potentially be addressed by slightly increasing the sealing bath depth so that the main body sits deeper in the bath. However, this may reintroduce clogging of the internal cavities, so different sealing bath depths should be further explored to find the optimum depth. Additionally,} 
shallow air pockets were introduced when the base lid was closed for curing; however, this does not affect functionality since these pockets are filled with silicone from the sealing layer. This might be addressed by dipping the base lid in uncured silicone before placing it, so there is no silicone-to-PLA interaction.

The presented robust and repeatable fabrication technique for pneumatic soft actuators addresses the clogging of internal cavities shown in Fig.~\ref{fig.cloggingRealWorld} and is a step towards the realization of soft robots as robust and scalable devices in real-world applications such as robotic rehabilitation, robotic surgery, and complex manufacturing processes.

%------------------------------------------------------------------
%supplementary video
%------------------------------------------------------------------
%certain highlighted sections of tutorial:
%	silicone/tutorials/sensorEmbedding/spectraFlexSensor/


%----------------------------------------------------------------
%\section*{APPENDIX}
%----------------------------------------------------------------

%----------------------------------------------------------------
%\section*{ACKNOWLEDGMENT}
%----------------------------------------------------------------
%We should thank the other members of the senior design team here.

%----------------------------------------------------------------
\bibliographystyle{unsrt}
\bibliography{references}
%----------------------------------------------------------------

\end{document}